\documentclass[a4paper,fleqn]{cas-dc}

\usepackage[numbers]{natbib}

\usepackage{multirow}
\usepackage{arydshln}
\usepackage{algorithm}
\usepackage{setspace}
\usepackage{algpseudocode}
\usepackage{dsfont}
\usepackage{bbm}

\def\tsc#1{\csdef{#1}{\textsc{\lowercase{#1}}\xspace}}
\tsc{WGM}
\tsc{QE}
\tsc{EP}
\tsc{PMS}
\tsc{BEC}
\tsc{DE}

\begin{document}
\let\WriteBookmarks\relax
\def\floatpagepagefraction{1}
\def\textpagefraction{.001}
\shorttitle{Leveraging social media news}
\shortauthors{CV Radhakrishnan et~al.}

\title [mode = title]{Independently Keypoint Learning for  Small Object Semantic Correspondence}

\author[1,2]{Hailong Jin}[orcid=0009-0009-6467-0779]

\author[1,2]{Huiying Li}[orcid=0000-0002-3637-2581]
\ead[url]{lihuiying@jlu.edu.cn}
\cormark[1]

\address[1]{Jilin University, College of Computer Science and Technology, Changchun, Jilin Province, China}
\address[2]{Key Laboratory of Symbol Computation and Knowledge Engineering, Jilin University, Changchun 130012, China}

\cortext[cor1]{Corresponding author}


\maketitle

\begin{abstract}
    Semantic correspondence remains a challenging task for establishing correspondences between a pair of images with the same category or similar scenes due to the large intra-class appearance. In this paper, we introduce a novel problem called 'Small Object Semantic Correspondence (SOSC).' This problem is challenging due to the close proximity of keypoints associated with small objects, which results in the fusion of these respective features. It is difficult to identify the corresponding key points of the fused features, and it is also difficult to be recognized. To address this challenge, we propose the Keypoint Bounding box-centered Cropping (KBC) method, which aims to increase the spatial separation between keypoints of small objects, thereby facilitating independent learning of these keypoints. The KBC method is seamlessly integrated into our proposed inference pipeline and can be easily incorporated into other methodologies, resulting in significant performance enhancements.  Additionally, we introduce a novel framework, named KBCNet, which serves as our baseline model. KBCNet comprises a Cross-Scale Feature Alignment (CSFA) module and an efficient 4D convolutional decoder. The CSFA module is designed to align multi-scale features, enriching keypoint representations by integrating fine-grained features and deep semantic features. Meanwhile, the 4D convolutional decoder, based on efficient 4D convolution, ensures efficiency and rapid convergence. To empirically validate the effectiveness of our proposed methodology, extensive experiments are conducted on three widely used benchmarks: PF-PASCAL, PF-WILLOW, and SPair-71k. Our KBC method demonstrates a substantial performance improvement of 7.5\% on the SPair-71K dataset, providing compelling evidence of its efficacy.
\end{abstract}

\section{Introduction}
\begin{sloppypar}

Semantic correspondence has been a fundamental problem in the field of computer vision, which focuses on identifying and establishing matching correlations between local features of image pairs containing the same category. It can facilitate many other computer vision tasks, such as semantic segmentation \cite{long2015fully, poudel2019fast, strudel2021segmenter}, object detection \cite{redmon2016you, yu2016unitbox, carion2020end}, optical flow \cite{ilg2017flownet, yang2019volumetric, xu2022gmflow} etc. Despite these promising applications, establishing semantic correspondence remains a challenging task due to large intra-class variations in appearance. In this paper, we introduce a novel challenge, \textit{Small Object Semantic Correspondence} (SOSC).

The advent of deep neural networks has revolutionized the field, allowing for the extraction of discriminative image features from pre-trained models like ResNet, eliminating the need for hand-crafted features\cite{lowe2004distinctive, bay2006surf}. Early deep learning approaches \cite{jeon2018parn, min2019hyperpixel, min2020learning} try to utilize multi-scale features to enhance the matching features, subsequently, some studies focus on designing effective correlation decoder \cite{rocco2018neighbourhood, cho2021cats, kim2022transformatcher}, e.g. 4D convolution decoder or transformer decoder, to refine score maps. Although these methods have achieved great success in semantic correspondence, they all face challenges in dealing with SOSC problem.

\begin{figure}
  \centering
  \includegraphics[scale=0.22]{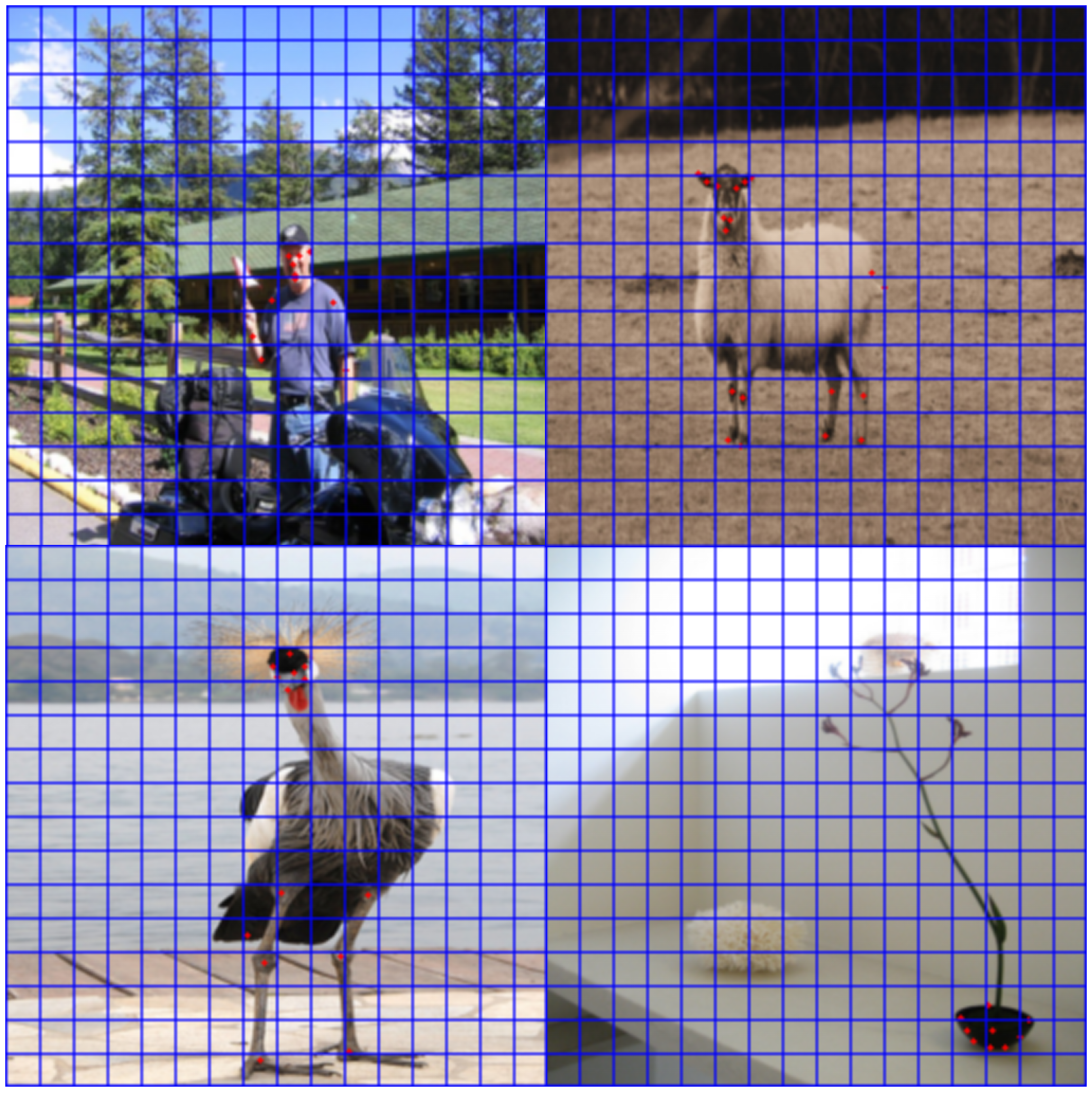}
  \caption{The visualization of input images for inference. The red points are the keypoints of the object, which can serve as either source or predicted target keypoints. The image is segmented into small regions using a $16\times 16$ window, typically aligned with the downsampling factor.}
  \label{fig:small_obj}
\end{figure}

Based on our analysis, The core aspect of this issue is the close proximity of keypoints associated with small objects. This spatial closeness often results in the merging of their deep layer features due to the downsampling module of the feature extractor, thereby making it difficult to identify their corresponding keypoints in the target image or to be identified. As depicted in Fig.~\ref{fig:small_obj}, there are several keypoints in a person's face within a $16\times 16$ window, after extracting features in 1/16 resolution, these keypoints will be merged into one inseparable vector. Similar examples can also be found in other categories.

In this paper, we propose a \textit{Keypoint Bounding box-centered Cropping} (KBC) method, which addresses the aforementioned issue from the image itself, and will not introduce any additional training module. To increase the separation between closely located keypoints, a direct and simple method is to resize the image to a larger size, however, training on a larger image size is costly and requires more computational resources. Considering the limited resources and more applicable in real applications, our KBC method can well handle this scenario. Specifically, through the positions of source keypoints, we can determine if this object is a small object according to the ratio of keypoint bounding box and target size. Once the image contains a small object, we will enlarge the image and crop it to the target size. In this way, we keep the input resolution unchanged while increasing the distance of closed keypoints.

Additionally, we propose a novel framework, called KBCNet, as our baseline model to validate the effectiveness of our proposed method. KBCNet consists of a \textit{Cross-Scale Feature Alignmen} (CSFA) module and an efficient 4D convolutional decoder. CSFA module takes the 1/16 resolution feature map as an anchor, and aggregates fine-grained features from 1/8 resolution and deep semantic features from 1/32 resolution. We take efficient 4D convolution as the basic module as our decoder for its fast convergence and efficiency.

Our main contributions can be summarized as follows:

\begin{itemize}
  \item We introduce a novel novel for semantic correspondence, the SOSC problem. To the best of our knowledge, we are the first to address this challenge. 
  \item We propose the KBC method to tackle the SOSC problem, serving as a plug-and-play module that can be seamlessly integrated with other methods without additional training.
  \item We design a novel framework, termed KBCNet, incorporating a CSFA module for leveraging cross-scale features and an efficient 4D convolutional decoder for enhancing local matching.
  \item We conduct extensive experiments to validate the effectiveness of our proposed KBC method on several public models. The results demonstrate substantial improvements in the SPair-71k dataset.
\end{itemize}

\end{sloppypar}

\section{Related Work}
\begin{sloppypar}
\textbf{Semantic Correspondence.} Semantic correspondence entails identifying the most relevant semantic matches between two images that share the same category or depict a similar scene.  Early approaches \cite{bristow2015dense, ham2016proposal} rely on hand-crafted features \cite{lowe2004distinctive, dalal2005histograms} for establishing correspondence in image pairs. Subsequent algorithms have transitioned to employing deep convolutional neural networks for the extraction of semantic features. Han et al. \cite{han2017scnet} propose an end-to-end trainable framework that matches the object proposals while maintaining their geometric consistency. Ufer and Ommer \cite{ufer2017deep} select salient features and then introduce a candidate-driven Markov Random Field (MRF) matching based on these discriminative features. To further enhance the precision of matching and filter out ambiguous results, Rocco et al. \cite{rocco2018neighbourhood} design a neighbourhood consensus neural network in 4D space. Zhao et al. \cite{zhao2021multi} introduce a multi-scale matching network that leverages multi-scale features to improve matching accuracy. In recent years, attention mechanisms have been employed to refine global matching. Kim et al. \cite{kim2022transformatcher} and Sun et al. \cite{sun2023correspondence} propose transformer-based architectures aimed at enabling global interactions between matches, further advancing the state-of-the-art in semantic correspondence.

\textbf{Small Object Problem.} Dealing with small object samples has posed a formidable challenge in the realm of computer vision, spanning various tasks such as object detection \cite{cao2019improved, zhao2019m2det}, semantic segmentation \cite{wang2018understanding, zhang2018context}, and image recognition \cite{zoph2018learning, dosovitskiy2020image}. These tasks mainly suffer from a lack of sufficient target-specific information caused by model downsampling. However, semantic correspondence faces a new challenge: the features of adjacent keypoint in small objects tend to blend together, rendering the keypoints inseparable. To address these two problems, we introduce the KBC, a simple yet effective approach.
\end{sloppypar}

\begin{figure*}
    \centering
    \includegraphics[scale=0.66]{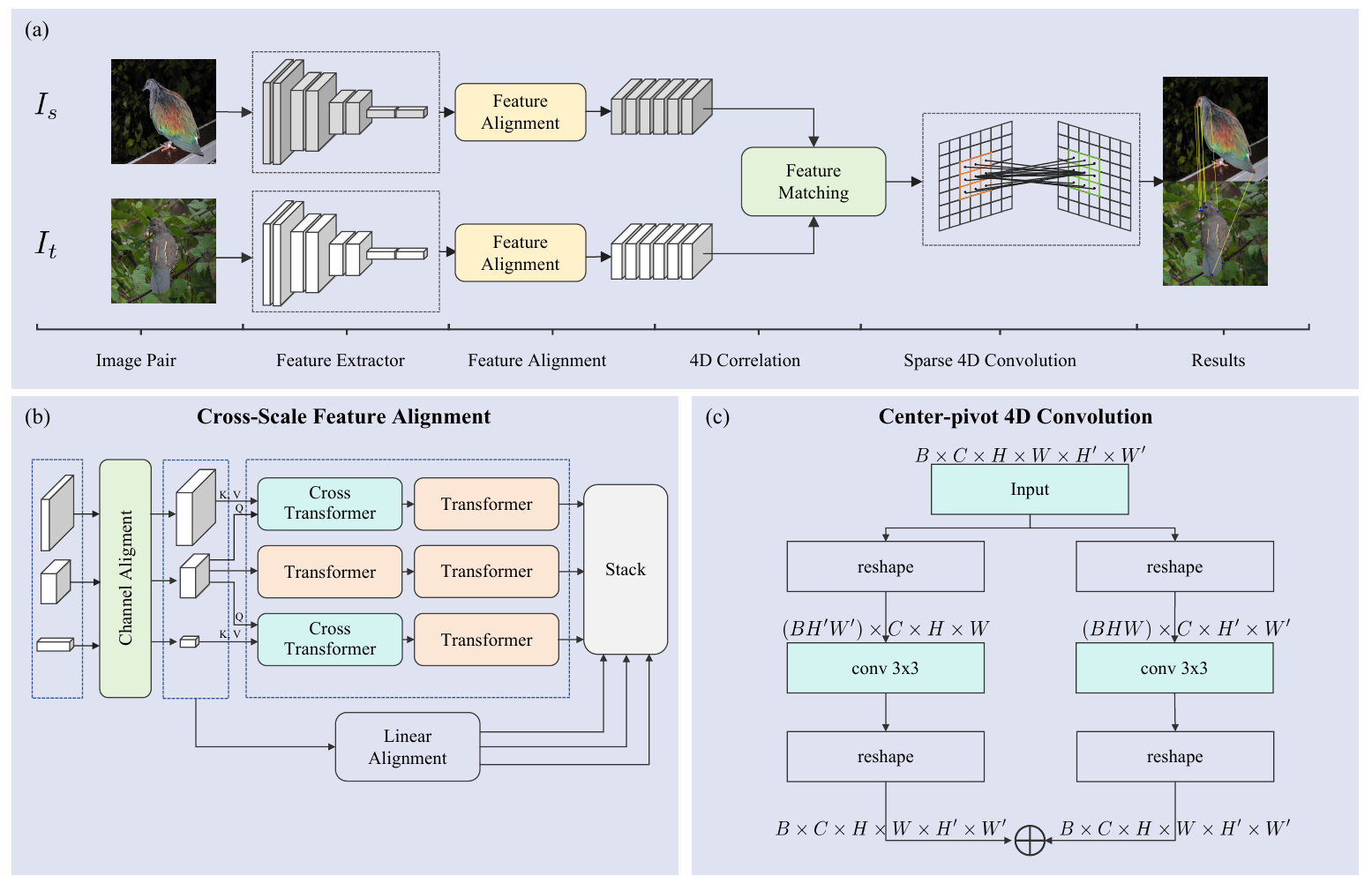}
    \caption{The overall architecture of the proposed framework. The framework comprises a convolutional backbone, a cross-scale feature alignment (CSFA) module, and an efficient 4D convolutional decoder. In the CSFA module, we use the second scale features as the query to integrate the other two scale features, respectively, and combine bilinear interpolation-aligned features. Subsequently, the aligned features are employed to compute 4D correlation maps with a similarity function. Finally, the obtained 4D correlation maps are input into the 4D convolutional decoder, which serves to adjust local matches.}
    \label{fig:architecture}
\end{figure*}

\textbf{Cross-Attention.} Cross-attention can capture relationships between image pairs, contributing to various computer vision tasks, such as depth estimation \cite{li2021revisiting}, optical flow \cite{shi2023flowformer++}, image matching \cite{jiang2021cotr} and style transfer \cite{wu2021styleformer}. Guizilini et al. \cite{guizilini2022multi} employ cross-attention for cost volume generation between image pairs, while Lu et al. \cite{lu2023transflow} leverage it for information exchange between consecutive frames. Cross-attention can also be used to transfer the source style into the required pose \cite{zhou2022cross}. Sun et al. \cite{sun2023correspondence} propose asymmetric feature learning through a combination of self-attention and cross-attention. Our work uniquely applies cross-attention to harness fine-grained and deep semantic features across scales within the same image.

\section{Proposed Method}

\begin{sloppypar}

When provided with two images, denoted as $I^s \in \mathbf{R}^{3 \times H \times W}$ and $I^t \in \mathbf{R}^{3 \times H' \times W'}$, with $I^s$ denoting the source image and $I^t$ denoting the target image. Our primary objective is to identify a set of matches denoted as $\{(p_i^s, p_i^t) | i = 1, ..., n\}$, where $p_i^s$ represents a keypoint in the source image, and $p_i^t$ corresponds to the matching target keypoint, with $n$ indicating the total number of keypoints that necessitate alignment between the two images.

In this section, we first introduce our proposed framework, KBCNet, followed by the presentation of our proposed KBC method that facilitates the independent learning of keypoints. As illustrated in Figure \ref{fig:architecture}, we extract a pair of image features based on a convolutional neural network, typically ResNet, pre-trained on the ImageNet \cite{deng2009imagenet} dataset. Subsequently, we pass these features to a CSFA module to align cross-scale features in the spatial dimension while integrating the fine-grained and deep semantic target information. The aligned features are then used to construct 4D correlation maps in the feature matching module. In the decoder part, we introduce an efficient 4D convolution module to refine matching relationships. Then we convert the predicted flow to the positions of matching points in the target image. During inference, we employ the proposed KBC method. This method involves enlarging the source image to increase the separation of closely located keypoints on small objects if required and then cropping it to match the input size based on the keypoint bounding box. Subsequently, we use the cropped source image to predict target keypoints. A similar operation is conducted for the target image based on estimated target keypoints, and finally both cropped source and target images are passed to the model for the final prediction. It is worth noting that the KBC method is only performed when dealing with small objects.

\subsection{Cross-Scale Feature Alignment and Matching}

To leverage cross-scale features and establish 4D correlation maps, aligning the features in the spatial dimension is crucial. Previous methods primarily adopt a bilinear interpolation method to align the image features. However, this method doesn't consider which information should receive more attention.

In this section, we propose an attention-based method to leverage cross-scale features. Given an image $I \in \mathbf{R}^{3 \times H \times W}$, we extract hierarchical image features and utilize only the output features from the last three building blocks, denoted as $F_1 \in \mathbf{R}^{ (\frac{H}{8} \times \frac{W}{8}) \times C}, F_2 \in \mathbf{R}^{(\frac{H}{16} \times \frac{W}{16}) \times 2C}, F_3 \in \mathbf{R}^{(\frac{H}{32} \times \frac{W}{32}) \times 4C}$, where $C$ is the channel dimension of $F_1$. First, we utilize linear layers to align the channel dimensions of $F_1$ and $F_3$ with $F_2$. Then, we utilize $F_2$ as the query to aggregate fine-grained target information from $F_1$ and deep semantic information from $F_3$.

\begin{align}
    F_{fine} &= CrossAttention(F_2, F_1, F_1) + F_2 \\
    F_{fine} &= MLP(F_{fine}) + F_{fine} \\
    F_{sem} &= CrossAttention(F_2, F_3, F_3) + F_2 \\
    F_{sem} &= MLP(F_{sem}) + F_{sem} 
\end{align}

Where $F_{fine}$ and $F_{sem}$ denote the features integrating fine-grained and deep semantic information, respectively. $MLP$ represents multi-layer perceptron, which contains two linear layers and a GELU activation function. $F_{fine}$ and $F_{sem}$ contains more target-specific information. Additionally, we employ self-attention to update $F_2$:

\begin{align}
    F_{self} &= SelfAttention(F_2) + F_2 \\
    F_{self} &= MLP(F_{self}) + F_{self}
\end{align}

Next, three self-attention layers are employed to refine $F_{fine}$, $F_{sem}$, and $F_{self}$ followed by the corresponding MLP layers. We use bilinear interpolation to align cross-scale features in the linear alignment module to enrich features, resulting in $F_1', F_2', F_3'$. Finally, we combine these features into a feature list, denoted as $\mathcal{F}$:

\begin{equation}
    \mathcal{F} = \{F_{fine}, F_{sem}, F_{self}, F_1', F_2', F_3' \} \label{eq:1}
\end{equation}

Based on Equation \ref{eq:1}, we can compute the feature map lists of the source image and the target image, denoted as $\mathcal{F}^s$, and $\mathcal{F}^t$, respectively. Then, the correlation map can be calculated as follows:

\begin{equation}
    Corr_i = \frac{\mathcal{F}_i^{s} \mathcal{F}_i^{tT}}{ \lVert \mathcal{F}_i^s \rVert \lVert \mathcal{F}_i^{tT}\rVert}
\end{equation}

Here, $\lVert \cdot \rVert$ denotes the Frobenius norm, $i$ represents the index of elements within $\mathcal{F}$. We compute the score maps for all corresponding features between $\mathcal{F}^s$ and $\mathcal{F}^t$, which are then stacked together to form $Corr \in \mathbf{R}^{6 \times (\frac{H}{16} \times \frac{W}{16}) \times (\frac{H'}{16} \times \frac{W'}{16})}$. Considering batch size, it is subsequently reshaped into a 4D correlation map $Corr \in \mathbf{R}^{B \times 6 \times \frac{H}{16} \times \frac{W}{16} \times \frac{H'}{16} \times \frac{W'}{16}}$.

\subsection{Sparse 4D Convolution}

In this paper, we reevaluate the potential of 4D convolution. In response to the high computational cost associated with traditional 4D convolution, we introduce a sparse 4D convolution \cite{min2021hypercorrelation}, known as center-pivot 4D convolution, as the basic component of our correlation decoder.

Given a 4D correlation map $M \in \mathbf{R}^{h \times w \times h' \times w'}$ and a 4D convolution kernel $k_{4D} \in \mathbf{R}^{k \times k \times k' \times k'}$. The traditional 4D convolution considers all matches between the tokens from a $k \times k$ local window in the source image and the tokens from a $k' \times k'$ local window in the target image, resulting in substantial computational requirements. However, center-pivot 4D convolution only considers significant matches. Assuming that the centers of local windows $k\times k$ and $k' \times k'$ are $x_c$ and $x_c'$, respectively. The center-pivot 4D convolution only considers matches relevant to these two center points. Therefore, the center-pivot 4D convolution can be formulated as:

\begin{equation}
    \begin{aligned}
    (M * k_{\text{CP}})(x_c, x_c') &= (M * k_c)(x_c, x_c') \\
                            & + (M * k_{c'})(x_c, x_c')
    \end{aligned}
\end{equation}

where $k_{\text{CP}}$ denotes the center-pivot 4D kernel, with $k_c \in \mathbf{R}^{k \times k \times 1 \times 1}$ and $k_{c'} \in \mathbf{R}^{1 \times 1 \times k' \times k'}$. It is noteworthy that $k_c$ and $k_{c'}$ can be further simplified to 2D kernels:

\begin{equation}
    \begin{aligned}
        (M * k_{\text{CP}})(x_c, x_c') = & \sum_{p' \in P(x_c')} f(x_c, p')k_{2D}^{'}(p'-x_c')  \\
                        &+ \sum_{p \in P(x_c)} f(p, x_c')k_{2D}(p-x_c) 
    \end{aligned}
\end{equation}

Here, $P(x_c)$ and $P(x_c')$ represent the neighbors of $x_c$ and $x_c'$, respectively. Both of $x_c$ and $x_c'$ are 2D coordinates. The 4D convolution therefore can be replaced by two 2D convolution layers, which greatly reduces the computational cost. We illustrate the entire calculation process of center-pivot 4D convolution in Figure \ref{fig:architecture} (c).


Our correlation decoder consists of three groups of sparse 4D convolution modules, each of which comprises the following components: a center-pivot 4D convolution, a group norm function, and a ReLU activation function. The final output of this decoder is linearly mapped to one dimension and then used to estimate the flow map and further calculate the location of the target keypoints.

\subsection{Training}

For the training objective, we follow the works of \cite{cho2021cats, sun2023correspondence}, which treat the semantic correspondence as a regression task. This is accomplished by utilizing the ground-truth keypoints of paired images to compute the training objective, referred to as the semantic flow map. To quantify the training loss, we calculate the average Euclidean distance between the predicted flow map and the ground-truth flow map, a metric known as the Average End-Point Error (AEPE) \cite{melekhov2019dgc} loss:

\begin{equation}
    \mathcal{L} = \lVert F_{pred} - F_{GT} \rVert_2
\end{equation}

Where $F_{pred}$ and $F_{GT}$ denote the predicted flow map and the ground-truth flow map, respectively.

\subsection{Inference Pipeline}
In our observations, we note that the keypoints of small objects exhibit close proximity, leading to the fusion of their features following downsampling. This phenomenon, in turn, causes predicted target points to cluster tightly, even when the actual locations are dispersed. We term this issue the SOSC problem. Existing models commonly employ downsampling to reduce model complexity and resource consumption. However, such a design introduces challenges in addressing the issue mentioned above. One straightforward strategy is to upscale the input image to a higher resolution, such as $512 \times 512$ \cite{hong2022integrative, sun2023correspondence}, but this comes at the expense of considerable computing resources.

To address the challenge of the SOSC problem, we introduce a keypoint bounding box-centered cropping (KBC) method into the inference pipeline. This method can seamlessly enhance the performance of existing methods, leading to substantial improvements in small object matching. 

We present the entire inference process and the details of the KBC method in Algorithm \ref{alg1}. Given a pair of images $I^s$ and $I^t$, along with their corresponding keypoints $P^s$ and $P^t$, the process begins by assessing whether the source image contains a small object. This determination is made by computing the ratio of the width and height of the keypoint bounding box to the input size: $r = max(w_{box} / w, h_{box}/h)$, where $w_{box}$ and $h_{box}$ are the width and height of the keypoint bounding box, while $(w, h)$ represent the input image size. If $r$ is lower than a specified threshold, we consider the image to contain a small object. The threshold is set below one since the keypoints need contextual information. If a small object is identified in the source image, the KBC method is applied to preprocess the image. Initially, we directly crop the image if the minimum distance between all keypoints exceeds 16; otherwise, we enlarge and crop it. Subsequently, we utilize the cropped source and target images to estimate the locations of the target keypoints. Similarly, we perform the same operation on the target image based on the estimated target keypoints.

\end{sloppypar}

\section{Experiments}

\begin{sloppypar}
\textbf{Datasets.} We conduct experiments on three public benchmarks: PF-PASCAL \cite{ham2017proposal}, PF-WILLOW \cite{ham2016proposal}, and SPair-71k \cite{min2019spair}. PF-PASCAL contains 1,351 pairs from 20 categories, and PF-WILLOW contains 900 image pairs from 4 categories. The PF-WILLOW dataset is only used to test models trained on PF-PASCAL. SPair-71k is a more challenging dataset that contains 70,958 image pairs of 18 categories. 
\end{sloppypar}

\textbf{Evaluation metric.} We adopt the percentage of correct keypoints (PCK) as our evaluation metric. Given the ground-truth keypoints $P$ and the predicted keypoints $P'$, PCK is measured by $PCK = \frac{1}{n}\sum_{i=1}^n \mathbbm{1}\{d_i < \alpha \cdot  max(h, w)\}$, where $d_i=\lVert P_i - P'_i \rVert$ is the Euclidean distance between the predicted point and ground-truth point, $h$ and $w$ are the height and width of an image or an object bounding box, respectively. Here, $\alpha$ denotes the threshold. Specifically, PF-PASCAL is evaluated using $\alpha_{\text{img}}$, SPair-71k and PF-WILLOW are evaluated using $\alpha_{\text{bbox}}$.

\begin{algorithm}
    \setstretch{1.2}
    \caption{Inference Pipeline}
    \begin{algorithmic}
    \Function{KBC}{$I, P$}
        \State $bbox \gets \text{getBoundingBox}(P)$
        \State $center \gets \text{getCenter}(bbox)$
        \State $min\_dis \gets \text{calculateMinimumDistance}(P)$
        \If {$min\_dis > 16$}
            \State $I_{crop}, P_{crop} \gets \text{centerCrop}(I, P, center)$
            \State \Return $I_{crop}, P_{crop}$
        \EndIf
        \State $I_{rsz}, P_{rsz} \gets \text{resizeImage}(I, P)$ 
        \State $bbox_{rsz} \gets \text{getBoundingBox}(P_{rsz})$
        \State $center_{rsz} \gets \text{getCenter}(bbox_{rsz})$
        \State $I_{crop}, P_{crop} \gets \text{centerCrop}(I_{rsz}, P_{rsz}, center_{rsz})$
        \State \Return $I_{crop}, P_{crop}$
    \EndFunction

    \Function{Inference}{$I^s, I^t, P^s$}
        \If {\text{containSmallObject}($P^s$)}
            \State $I^s, P^s \gets$ \text{KBC}($I^s, P^s$)
        \EndIf
        \State $F_{pred} \gets \text{network}(I^s, I^t)$
        \State $P^t_{pred} \gets \text{getKeypoints}(F_{pred}, P^s)$
        \If {\text{containSmallObject}($P^t_{pred}$)}
            \State $I^t, \_ \gets$ \text{KBC}($I^t, P^t_{pred}$)
            \State $F_{pred} \gets \text{network}(I^s, I^t)$
            \State $P^t_{pred} \gets \text{getKeypoints}(F_{pred}, P^s)$
        \EndIf
        \State \Return $P^t_{pred}$
    \EndFunction
    \end{algorithmic}
    \label{alg1}
\end{algorithm}

\begin{sloppypar}
\textbf{Implementation Details.} We leverage the ResNet-101 pre-trained on ImageNet \cite{deng2009imagenet} to extract image features, and utilize the outputs derived from conv3\_x, conv4\_x, and conv5\_x. The input resolution for the source and target images is $256 \times 256$. Our model is implemented in Pytorch and trained on a single NVIDIA GeForce RTX4090 GPU. During training, we use an Adam optimizer with a learning rate of 1e-5. We train the model for 100 epochs on the PF-PASCAL dataset and 10 epochs on the SPair-71k dataset. The batch size is set to 16 for both datasets. In the inference stage, we set the thresholds to 0.7, 0.9, and 0.8 for PF-PASCAL, PF-WILLOW, and SPair-71k, respectively. For SPair-71k and PF-WILLOW datasets, if the target image is cropped, we project the predicted target positions back to the original setting to compute the PCK, ensuring a fair comparison.
\end{sloppypar}

\begin{table*}
    \centering
    \renewcommand\arraystretch{1.0}
    \caption{Comparison between our KBCNet and previous works on PF-Pascal, PF-Willow, and SPair-71K datasets. All the methods in the table utilize ResNet101 as the backbone to extract image features and are trained by strongly supervised loss. KBCNet$_{baseline}$ means the KBC method is not employed in the inference stage. The best results are in bold, and the second-best results are underlined.}
    \label{tab:1}
    \scalebox{1.0}{\begin{tabular}{l  |c | c  c  c  c c c c}
        \toprule
        \multirow{3}{*}{Methods} & \multirow{3}{*}{Aggregation} & SPair-71K & \multicolumn{3}{c}{PF-PASCAL} & \multicolumn{3}{c}{PF-WILLOW}  \\
        &  & PCK@$\alpha_{\text{bbox}}$ & \multicolumn{3}{c}{PCK@$\alpha_{\text{img}}$} & \multicolumn{3}{c}{PCK@$\alpha_{\text{bbox}}$}  \\
        & &  0.1 & 0.05 & 0.1 & 0.15 & 0.05 & 0.1 & 0.15 \\
        \midrule
        SCNet \cite{han2017scnet} &  - & - & 36.2 & 72.2 & 82.0 & 38.6 & 70.4 & 85.3 \\
        NCNet \cite{rocco2018neighbourhood} &  4D Conv. & 20.1 & 54.3 & 78.9 & 86.0 & 33.8 &67.0 & 83.7\\
        DCCNet \cite{huang2019dynamic} &  4D Conv. & - & 55.6 & 82.3 & 90.5 & 43.6 & 73.8 & 86.5 \\
        HPF \cite{min2019hyperpixel} &  RHM & 28.2 & 60.1 & 84.8 & 92.7 & 45.9 & 74.4 & 85.6\\
        SCOT \cite{liu2020semantic}  &  OT-RHM & 35.6 & 63.1 & 85.4 & 92.7 & 47.8 & 76.0 & 87.1\\
        DHPF \cite{min2020learning} &  RHM  &  37.3 & 75.7 & 90.7 & 95.0 & 49.5 & 77.6 & 89.1 \\
        ANCNet \cite{li2020correspondence} &  4D Conv. & - & - & 86.1 & - & - & - & - \\
        CHM \cite{min2021convolutional} &  6D Conv. &  46.3 & 80.1 & 91.6 & 94.9 & 52.7 & 79.4 & 87.5 \\
        MMNet-FCN \cite{zhao2021multi}  &  2D Conv.&  50.4 & \textbf{81.1} & 91.6 & 95.9 & - & - & - \\
        CATs \cite{cho2021cats}  & Transformer&  49.9 & 75.4 & 92.6 & 96.4 & 50.3 & 79.2 & 90.3\\
        TransforMatcher \cite{kim2022transformatcher} & Transformer &  \underline{53.7} & \underline{80.8} & 91.8 & - & - & 76.0 & - \\
        \midrule
        KBCNet$_{baseline}$ &  4D Conv.&   51.6 & 78.0 & \underline{93.6} & \underline{96.8} & \underline{54.0} & \underline{82.5} &  \underline{92.5}  \\
        \textbf{KBCNet} & 4D Conv. & \textbf{59.1} & 78.1 & \textbf{93.8} & \textbf{97.2} & \textbf{56.4} & \textbf{84.7} & \textbf{93.5}  \\
        \bottomrule
    \end{tabular}
    }
\end{table*}

\begin{table*}
\centering
\renewcommand\arraystretch{1.3}
\caption{Per-class evaluation on SPair-71k dataset. The results are under the setting $\alpha=0.1$. The results of SCOT , DHPF, CHM, CATs, and MMNet-FCN are come from their released pre-trained models. The best results are in bold.}
    \label{tab:2}
\scalebox{0.70}{
    \begin{tabular}{l|cccccccccccccccccc|c}
        \toprule
        Methods & aero. & bike & bird & boat & bott. & bus & car & cat & chai. & cow & dog & hors. & mbik. & pers. & plan. & shee. & trai. & tv & all \\
        \midrule
        CNNGeo~\cite{rocco2017convolutional} &  23.4 & 16.7 & 40.2 & 14.3 & 36.4 & 27.7 & 26.0 & 32.7 & 12.7 & 27.4 & 22.8 & 13.7 & 20.9 & 21.0 & 17.5 & 10.2 & 30.8 & 34.1 & 20.6  \\
    
        A2Net~\cite{seo2018attentive} & 22.6 & {18.5} & 42.0 & {16.4} & {37.9} & {30.8} & {26.5} & 35.6 & 13.3 & 29.6 & 24.3 & 16.0 & 21.6 & {22.8} & {20.5} & 13.5 & 31.4 & {36.5} & 22.3 \\

        NC-Net~\cite{rocco2018neighbourhood} & 17.9 & 12.2 & 32.1 & 11.7 & 29.0 & 19.9 & 16.1 & 39.2 & 9.9 & 23.9 & 18.8 & 15.7 & 17.4 & 15.9 & 14.8 & 9.6 & 24.2 & 31.1 & 20.1   \\

        HPF~\cite{min2019hyperpixel} & {25.2} & {18.9} & {52.1} & {15.7} & {38.0} & {22.8} & {19.1} & {52.9} & {17.9} & {33.0} & {32.8} & {20.6} & {24.4} & {27.9} & 21.1 & {15.9} & {31.5} & {35.6} & {28.2} \\

        \midrule

        {SCOT} \cite{liu2020semantic} & {34.8} & {20.3} & {64.2} & {21.0} & {43.3} & {26.8} & {21.7} & {63.6} & {19.6} & {42.6} & {42.2} & {30.9} & {29.2} & {35.5} & {28.1} & {24.3} & {47.3} & {40.3} & {35.5} \\
        
        \textbf{SCOT-KBC} & 38.0 & 25.7 & 67.4 & 24.6 & 45.7 & 29.8 & 22.8 & 63.8 & 22.5 & 51.5 & 44.8 & 36.5 & 31.7 & 40.7 & 32.2 & 30.7 & 47.7 & 47.7 & 39.2 \\

        \hdashline

        {DHPF} \cite{min2020learning} & {39.2} & {23.9} & {68.7} & {19.6} & {42.9} & {28.2} & {21.1} & {62.0} & {22.7} & {47.9} & {46.1} & {33.4} & {27.7} & {40.6} & {28.6} & {28.8} & {51.0} & {47.5} & {37.9} \\
        \textbf{DHPF-KBC} \cite{min2020learning} & {40.6} & {27.5} & {70.4} & {22.6} & {44.6} & {32.3} & {24.4} & {62.4} & {23.7} & {55.9} & {47.4} & {36.5} & {29.5} & {44.6} & {38.1} & {34.0} & {53.2} & {53.5} & {41.3} \\

        \hdashline
        
        {CHM}~\cite{min2021convolutional} & {49.6} & {29.3} & {68.7} & {29.7} & {45.3} & {48.4} & {39.5} & {64.8} & {20.3} & {60.6} & {56.1} & {45.9} & {33.8} & {44.3} & {39.0} & {31.3} & {72.2} & {55.6} & {46.4} \\
    
        \textbf{CHM-KBC} & {51.0} & {32.3} & {73.9} & {32.5} & {46.5} & {50.2} & {37.9} & {68.0} & {23.2} & {68.0} & {58.3} & {49.9} & {36.0} & {50.5} & {49.0} & {43.8} & {71.5} & {58.8} & {50.2} \\

        \hdashline

        CATs \cite{cho2021cats}  & {52.0} & {34.7} & {72.2} & {34.2} & {49.9} & {57.5} & {43.6} & {66.4} & {24.4} & {63.2} & {56.6} & {51.9} & {42.6} & {41.3} & {43.1} & {33.7} & {72.5} & {58.0} & {49.9} \\

        \textbf{CATs-KBC}  & {53.7} & {44.6} & {78.9} & {36.9} & {51.7} & {60.9} & {45.8} & {70.7} & {29.5} & {\textbf{74.0}} & {62.6} & {58.8} & {48.2} & {56.2} & {57.0} & {46.2} & {76.2} & {71.2} & {56.9} \\

        \hdashline

        MMNet-FCN \cite{zhao2021multi} & 57.0 & 37.0 & 66.3 & 35.8  & 48.9 & 63.4 & 50.1 & 66.6 & 31.2 & 70.6 & 56.8 & 53.2 & 41.7 & 35.7 & 40.8 & 36.7 & 81.5 & 68.0 & 52.2 \\

        \textbf{MMNet-FCN-KBC} & \textbf{57.0} & 38.5 & 71.4 & 37.6  & 48.9 & \textbf{66.1} & \textbf{51.5} & 69.2 & \textbf{32.3} & 73.0 & 58.1 & 56.1 & 46.3 & 45.7 & 55.6 & \textbf{47.5} & \textbf{81.9} & 72.8 & 56.1 \\
        
        \midrule
        KBCNet$_{baseline}$ & 52.0 & 36.7 & 73.4 & 34.5 & 53.8 & 59.1 & 44.5 & 65.6 & 26.0 & 61.4 & 59.1 & 54.3 & 44.4 & 45.6 & 41.1 & 32.0 & 75.5 & 69.5 & 51.6 \\
        \textbf{KBCNet} & 55.6 & \textbf{48.2} & \textbf{77.1} & \textbf{38.0} & \textbf{56.4} & 62.3 & 49.7 & \textbf{69.2} & 31.1 & 73.7 & \textbf{64.7} & \textbf{63.0} & \textbf{52.7} & \textbf{58.1} & \textbf{59.5} & 46.3 & 81.3 & \textbf{74.2} & \textbf{59.1} \\
        \bottomrule
    \end{tabular}
    }

\end{table*}

\begin{figure*}
    \centering
    \includegraphics[scale=0.65]{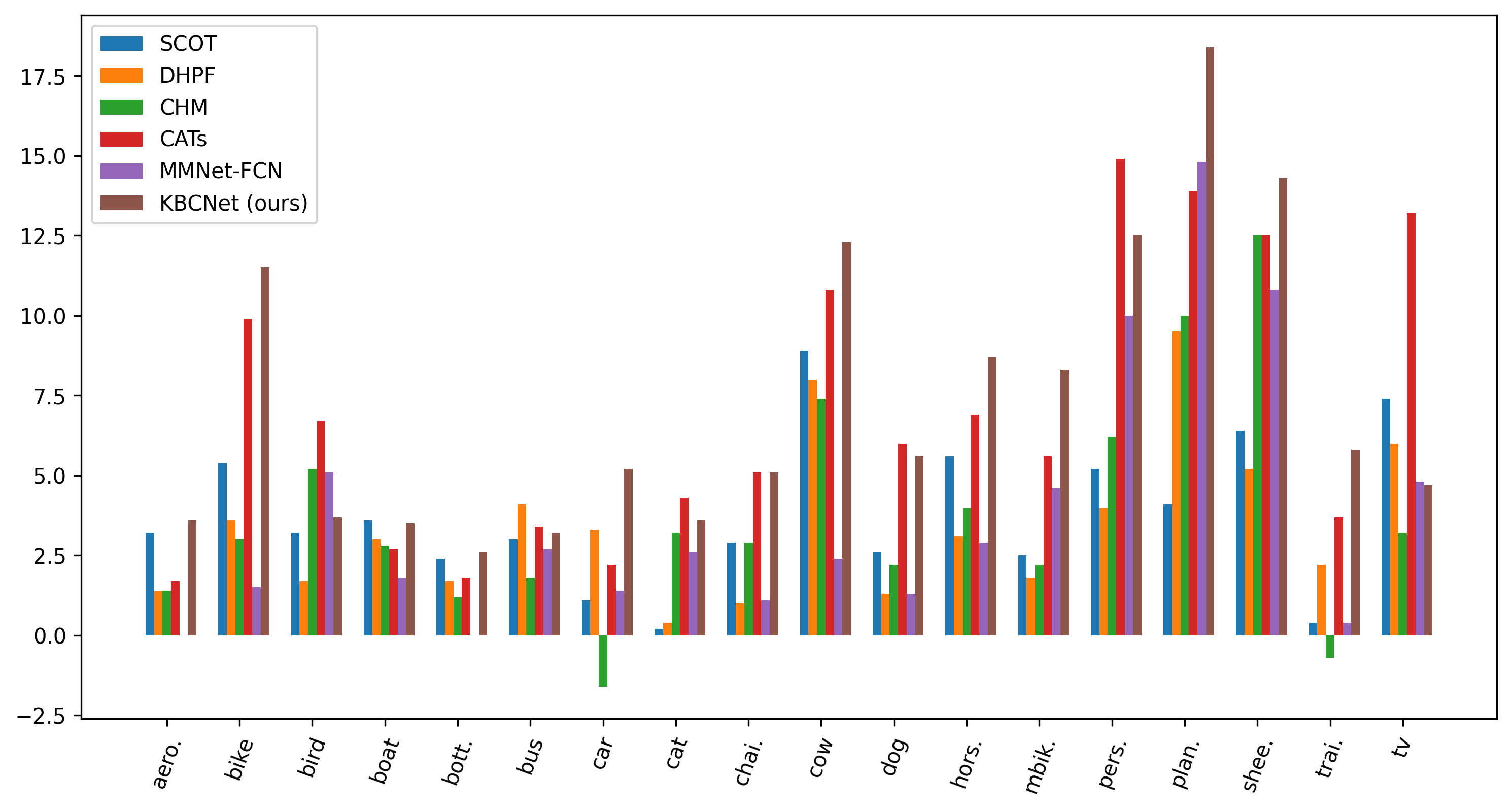}
    \caption{Visualization of the improvements our KBC method on SCOT, DHPF, CHM, CATs, MMNet-FCN, as well as our KBCNet on SPair-71k dataset.}
    \label{fig:differ}
\end{figure*}

\subsection{Results and analysis}

\begin{sloppypar}
We conduct a comprehensive analysis of our proposed method in comparison with other state-of-the-art models. The results of these comparisons are assessed on three widely recognized public benchmarks: SPair-71k \cite{min2019spair}, PF-PASCAL \cite{ham2017proposal}, and PF-WILLOW \cite{ham2016proposal}, as detailed in Table \ref{tab:1}. Our KBCNet achieves state-of-the-art performance in terms of PCK@0.1. Precisely, we attain an accuracy of 59.1\% on SPair-71k, 93.8\% on PF-PASCAL, and 84.7\% on PF-WILLOW. Our proposed method outperforms existing state-of-the-art models by 5.4\%, 1.2\%, and 5.3\% on PCK@0.1 in these three benchmark datasets. Additionally, our proposed baseline method also demonstrates superior performance to other methods on the PF-PASCAL and PF-WILLOW datasets under the setting of $\alpha = 0.1$. The results demonstrate the effectiveness of our KBCNet.
\end{sloppypar}

\begin{sloppypar}
Incorporating the KBC method enhances our proposed baseline model, yielding improvements of 7.5\%, 0.2\%, and 2.2\% on the SPair-71k, PF-PASCAL, and PF-WILLOW datasets, respectively. The significant improvements in the SPair-71k dataset can be attributed to the presence of numerous images containing small objects, which are effectively addressed and matched by our method. In contrast, we observe only marginal improvements in the PF-PASCAL dataset, as small objects are not a predominant limitation in this dataset.

To assess the generalization capability of our proposed KBC method, we evaluate its performance across five publicly available models: SCOT \cite{liu2020semantic}, DHPF \cite{min2020learning}, CHM\cite{min2021convolutional}, CATs \cite{cho2021cats}, and MMNet-FCN \cite{zhao2021multi}. As presented in Table \ref{tab:2}, our KBC method demonstrates significant improvements, achieving average increases of 3.7\%, 3.4\%, 3.8\%, 7.0\%, and 3.9\% for these models, respectively. Notably, CATs-KBC showcases notable enhancements of 14.9\%, 13.9\%, and 12.5\% for the categories of person, potted plant, and sheep. To provide a more intuitive understanding of the benefits brought by our method, we visualize the improvements from Table~\ref{tab:1} in Figure~\ref{fig:differ}. It is observed that the CHM model demonstrates less favorable performance in the car and train classes, suggesting a potential requirement for additional contextual information for these specific objects.

\subsection{Ablation Studies}

To comprehensively explore the optimal configuration and thoroughly evaluate the effectiveness of our proposed module, we undertake an extensive array of experiments.

\begin{table}
    \centering
    \caption{Ablation studies with and with out CSFA module on PF-PASCAL and SPair-71k.}
    \label{tab:3}
    \begin{tabular}{ c c c c}
        \toprule
        Methods & $\alpha$  & KBCNet (w/o CSFA) & KBCNet \\ 
        \midrule
        \multirow{1}{*}{Spair-71k} & 0.10 & 49.7 & \textbf{51.6}  \\
        \midrule
        \multirow{3}{*}{PF-PASCAL} & 0.05 & \textbf{79.0}  & 78.0 \\
        & 0.10 & 93.2  & \textbf{93.6} \\
        & 0.15 & 96.1 &  \textbf{96.8} \\
        \bottomrule
    \end{tabular}
\end{table}

\begin{table}
    \centering
    \caption{Ablation studies of the KBC method. KBC(src) and KBC(trg) represent applying the KBC method to the source image and target image, respectively. KBC(src+trg) means using the KBC method on both the source image and the target image.}
    \label{tab:5}
    \begin{tabular}{ l c c c}
        \toprule
        \multirow{3}{*}{Methods} & \multicolumn{3}{c}{SPair-71k} \\
        & \multicolumn{3}{c}{PCK@$\alpha_{bbox}$} \\
        & 0.05 & 0.1 & 0.15 \\
        \midrule
        KBCNet$_{baseline}$ & 30.0 & 51.6 & 63.1 \\
        \midrule 
        + KBC(src) & 32.6 & 54.9 & 65.6 \\
        + KBC(trg) & 35.5 & 55.3 & 65.2 \\
        + KBC(src+trg) & 39.1 & 59.1 & 68.2 \\
        \bottomrule
    \end{tabular}
\end{table}

\textbf{Effects of cross-scale feature alignment module.} To verify the effectiveness of our proposed CSFA module,  we conduct experiments comparing model performance with and without its inclusion. The empirical results, shown in Table \ref{tab:3}, clearly illustrate the improvements attributed to the integration of the CSFA module into our model architecture. Specifically, under the experimental setting with $\alpha=0.1$, our model equipped with the CSFA module demonstrates superior performance, achieving an enhancement of 1.9\% on the SPair-71k dataset and 0.4\% improvement on the PF-PASCAL dataset compared to the counterpart lacking the CSFA module.

\begin{table}
    \centering
    \caption{Comparison of center-pivot 4D convolution with traditional convolution in terms of memory, training time, inference time, and PCK value.}
    \label{tab:4}
    \scalebox{1.0}{
        \begin{tabular}{ l c c}
            \toprule
            \multirow{2}{*}{Methods} & center-pivot & traditional \\
            & 4D convolution  & 4D convolution \\
            \midrule
            Memory (GB) & 3.2 & 3.3 \\
            Training Time (min)  & 25 & 120 \\
            Infer Time (ms) & 30  & 61 \\
            PCK@$\alpha_{bbox}$ & 51.6 & 52.2 \\
            \bottomrule
        \end{tabular}
    }
    
\end{table}

\textbf{Effects of KBC method.} As illustrated in Table~\ref{tab:5}, we investigate the impact of KBC method when applied separately to source and target images, as well as in combination. Our KBC method yields an improvement of 3.3\% and 3.7\% when employing KBC method to source and target images, respectively, which is a large margin of gains. This observation suggests that both the source and target images encompass a portion of small objects. It is worth noting that if applying KBC method to both source and target images, we can obtain a significant improvements of 7.5\%, which demonstrates the application value of our KBC method.

\textbf{Analysis on 4D convolution.} To compare the performance of center-pivot 4D convolution with traditional 4D convolution, we report the results of memory usage, training time, inference time, and PCK value on the SPair-71k dataset as shown in Table \ref{tab:4}. Center-pivot 4D convolution has 4.8x faster training time and 2x faster inference time than traditional 4D convolution. Center-pivot 4D convolution sacrifices some accuracy to obtain the above advantages, which is acceptable.

\begin{figure}
    \centering
    \includegraphics[scale=0.55]{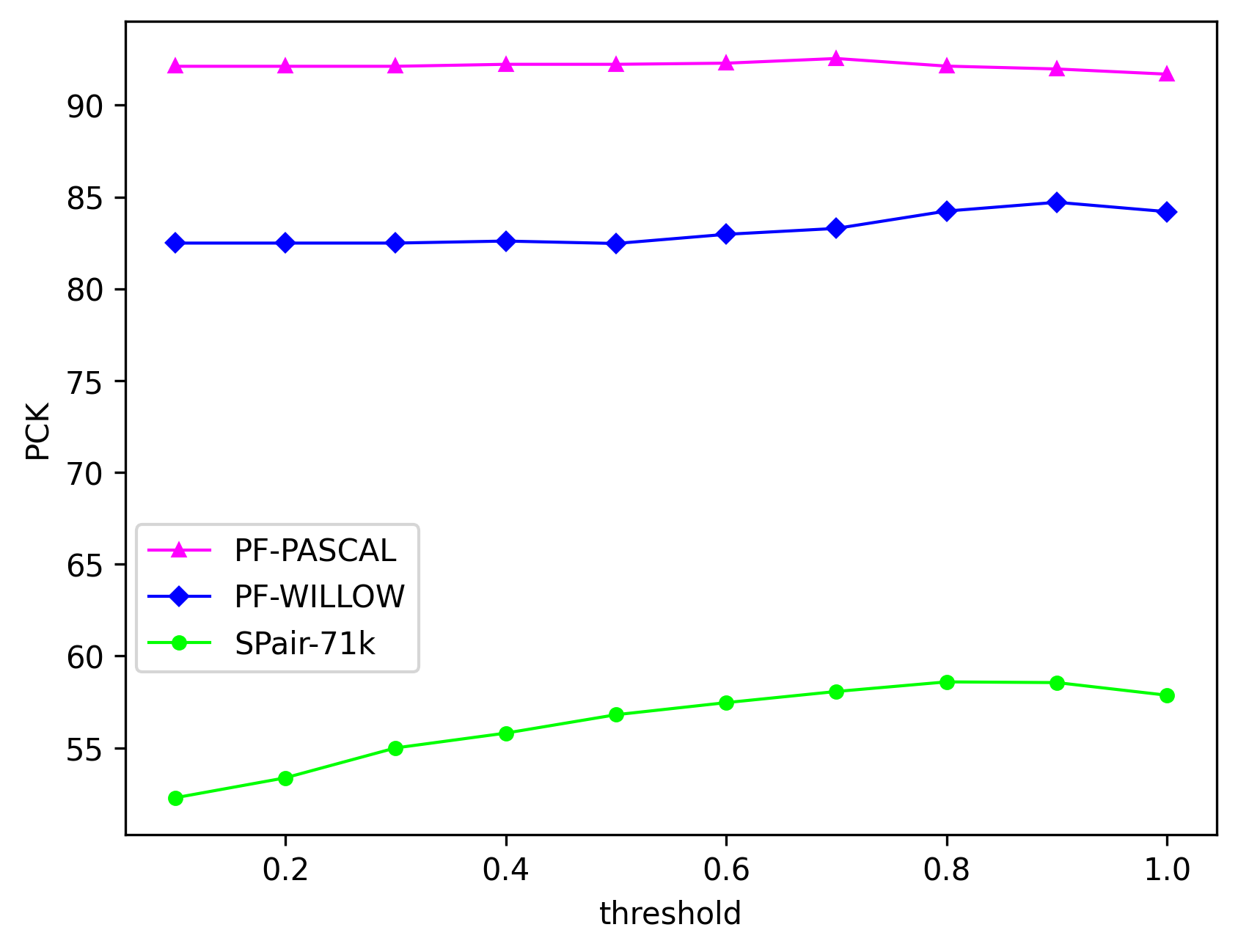}
    \caption{The threshold curves on PF-PASCAL, PF-WILLOW, and SPair-71k.}
    \label{fig:thres}
\end{figure}

\textbf{Effects of threshold.} To validate the impact of the threshold on the PF-PASCAL, PF-WILLOW, and SPair-71k datasets, we depict the curves corresponding to different thresholds in Figure \ref{fig:thres}. The results for PF-PASCAL and SPair-71k are derived from the validation dataset, while those for PF-WILLOW are obtained from the test dataset. Analyzing Figure \ref{fig:thres}, it is evident that our proposed KBC method exhibits a more pronounced effect on SPair-71k and PF-WILLOW, with a slight impact on PF-PASCAL. Optimal results are achieved by setting the threshold to 0.7, 0.9, and 0.8 for PF-PASCAL, PF-WILLOW, and SPair-71k, respectively.

\begin{figure*}
    \centering
    \includegraphics[scale=0.36]{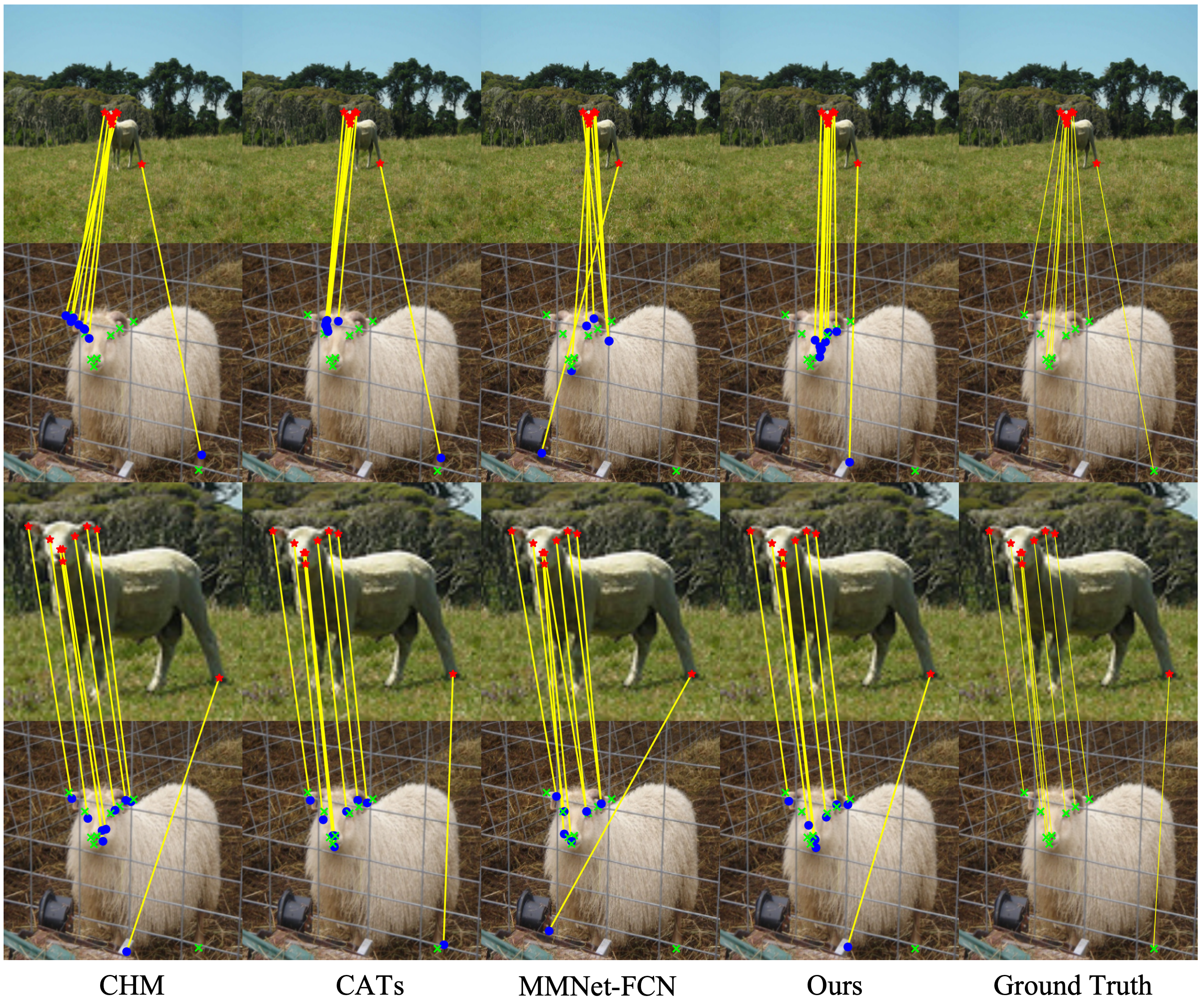}
    \caption{Small object visual correspondence generated by state-of-the-art methods, including CHM, CATs, MMNet-FCN and our proposed method. The first two rows of images represent the original matching results of these methods, and the last two rows represent the matching results using the KBC method. }
    \label{fig:vis}
\end{figure*}

\begin{figure*}
    \centering
    \includegraphics[scale=0.32]{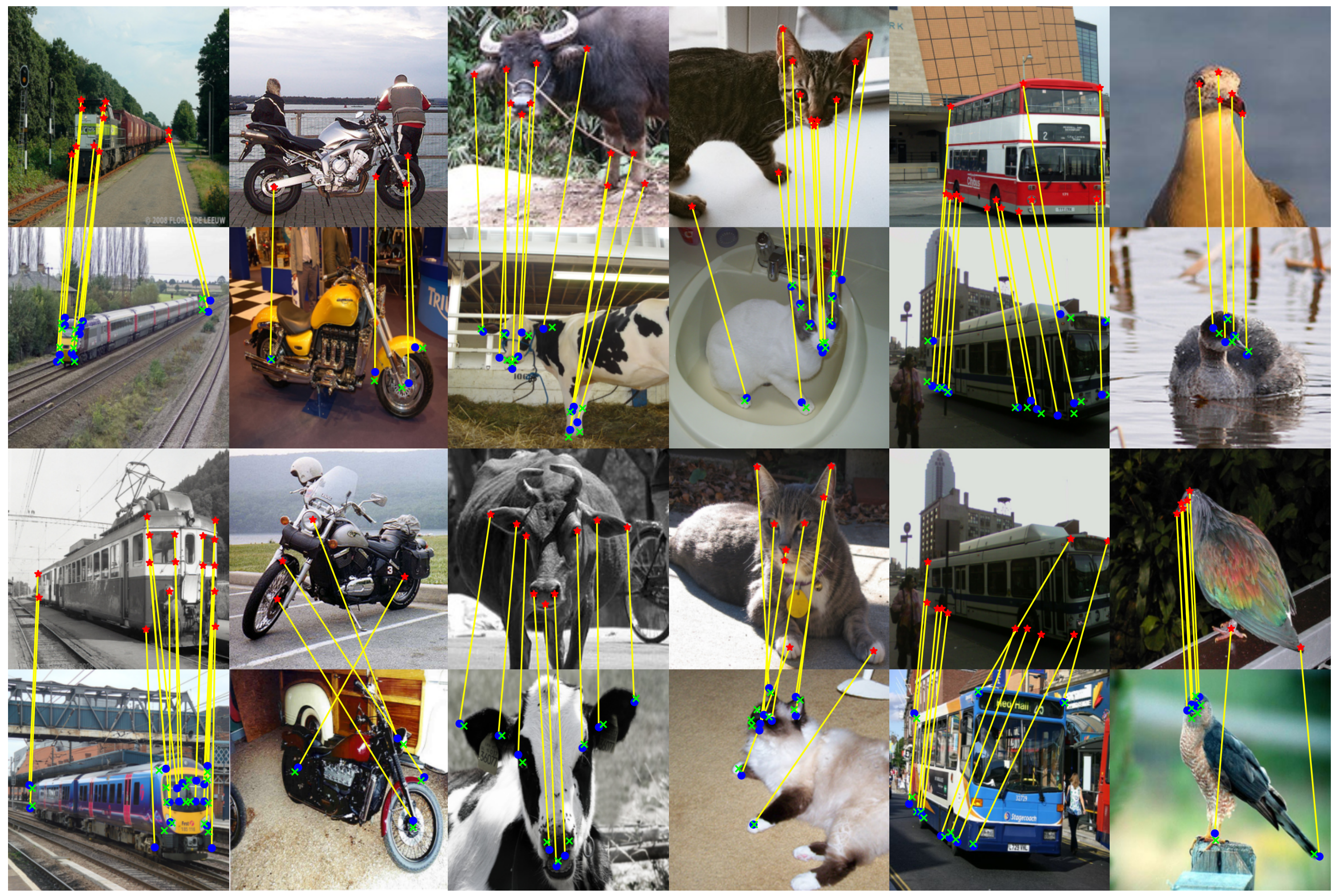}
    \caption{More matching visualization results for our KBCNet. The red star indicates the location of the source keypoint, the blue dot indicates the predicted target keypoint location, and the green "x" indicates the location of the ground truth target keypoint.}
    \label{fig:vis2}
\end{figure*}

\subsection{Qualitative Results and Visualization}

To conduct a qualitative analysis of our proposed KBC method, we visualize the matching results with and without the KBC method on three public models and our proposed method. As depicted in Figure \ref{fig:vis}, we can observe that the predicted target points of small objects with closely located source keypoints tend to appear close to each other as these source pixel features are fused after downsampling. Our proposed KBC method efficiently increases the spatial separation between these points, enabling them to be independently learned. Furthermore, our method can filter out extraneous background clutter, augmenting the signal-to-noise ratio and enhancing the effective target information. Besides, our method can be generalized to other methods without further training. More visualization results of our KBCNet are depicted in Figure~\ref{fig:vis2}.

\end{sloppypar}
\section{Conclusion}

\begin{sloppypar}
In this paper, we introduce the 'Small Object Semantic Correspondence (SOSC)' problem, which particularly emphasizes the difficulty posed by small object keypoints' close proximity and subsequent fusion of features. To mitigate this, we propose the Keypoint Bounding box-centered Cropping (KBC) method, which strategically augments spatial separation between keypoints, facilitating independent learning and identification. Integrated seamlessly into our proposed inference pipeline and adaptable to other methodologies. Moreover, we introduce the KBCNet framework as our baseline model, comprising a Cross-Scale Feature Alignment (CSFA) module and an efficient 4D convolutional decoder. The CSFA module enriches keypoint representations by aligning multi-scale features, while the 4D convolutional decoder ensures efficiency and rapid convergence. Empirical validation on established benchmarks, PF-PASCAL, PF-WILLOW, and SPair-71k demonstrates the efficacy of our approach. Notably, our KBC method exhibits a substantial 7.5\% performance improvement on the SPair-71k dataset, underscoring its effectiveness in addressing the challenges of semantic correspondence, particularly in scenarios involving small objects.
\end{sloppypar}

\bibliographystyle{cas-model2-names}
\bibliography{main}

\end{document}